\documentclass[conference]{IEEEtran}
\IEEEoverridecommandlockouts

\usepackage{cite}
\usepackage{amsmath,amssymb,amsfonts}
\usepackage{algorithmic}
\usepackage{graphicx}
\usepackage{textcomp}
\usepackage{xcolor}
\def\BibTeX{{\rm B\kern-.05em{\sc i\kern-.025em b}\kern-.08em
    T\kern-.1667em\lower.7ex\hbox{E}\kern-.125emX}}

\usepackage{epsfig}
\usepackage{graphicx}
\usepackage{subcaption}
\usepackage{amsmath}
\usepackage{amssymb}
\usepackage{booktabs}
\usepackage{xspace}
\usepackage{enumitem}
\usepackage{url}
\usepackage{hyperref}

\usepackage[bb=boondox]{mathalfa}

\usepackage[ruled,vlined]{algorithm2e}
\definecolor{commentcolor}{RGB}{110,154,155}

\newcommand{\PyComment}[1]{\ttfamily\textcolor{commentcolor}{\# #1}}
\newcommand{\PyNormal}[1]{\ttfamily\textcolor{commentcolor}{ #1}} 

\newcommand{\PyNormalTab}[1]{\ttfamily\textcolor{commentcolor}{\quad #1}}

\newcommand{\PyCode}[1]{\ttfamily\textcolor{black}{#1}}

\usepackage{verbatim}  

\makeatletter
\DeclareRobustCommand\onedot{\futurelet\@let@token\@onedot}
\def\@onedot{\ifx\@let@token.\else.\null\fi\xspace}
\def\eg{\emph{e.g}\onedot} 
\def\ie{\emph{i.e}\onedot}

\def\etal{\emph{et al}\onedot}
\makeatother



\definecolor{sky}{RGB}{0, 230, 230}
\definecolor{pinkcolor}{RGB}{239, 66, 245}
\definecolor{pinkcolor2}{RGB}{252,66,162}
\definecolor{purplecolor}{RGB}{88, 68, 208}
\definecolor{greencolor}{RGB}{50,205,50}
\definecolor{darkgreencolor}{RGB}{0,128,0}
\definecolor{orangecolor}{RGB}{242, 136, 34}
\definecolor{redcolor}{RGB}{215,25,28}
\begin{document}

\title{Leveraging Large Language Models for Suicide Detection on Social Media with Limited Labels\\
}

\author{\IEEEauthorblockN{Vy Nguyen}
\IEEEauthorblockA{
\textit{Northeastern University}\\
Boston, MA, USA \\
nguyen.vy7@northeastern.edu}
\and
\IEEEauthorblockN{Chau Pham}
\IEEEauthorblockA{
\textit{Boston University}\\
Boston, MA, USA  \\
chaupham@bu.edu}

}

\maketitle
\thispagestyle{plain} 
\pagestyle{plain} 

\begin{abstract}
The increasing frequency of suicidal thoughts highlights the importance of early detection and intervention. Social media platforms, where users often share personal experiences and seek help, could be utilized to identify individuals at risk. However, the large volume of daily posts makes manual review impractical. This paper explores the use of Large Language Models (LLMs) to automatically detect suicidal content in text-based social media posts. We propose a novel method for generating pseudo-labels for unlabeled data by prompting LLMs, along with traditional classification fine-tuning techniques to enhance label accuracy.
To create a strong suicide detection model, we develop an ensemble approach involving prompting with \textit{Qwen2-72B-Instruct}, and using fine-tuned models such as \textit{Llama3-8B}, \textit{Llama3.1-8B}, and \textit{Gemma2-9B}. We evaluate our approach on the dataset of the Suicide Ideation Detection on Social Media Challenge, a track of the IEEE Big Data 2024 Big Data Cup. Additionally, we conduct a comprehensive analysis to assess the impact of different models and fine-tuning strategies on detection performance. Experimental results show that the ensemble model significantly improves the detection accuracy, by $\mathbf{5}$\% points compared with the individual models. It achieves a weight F1 score of $\mathbf{0.770}$ on the public test set, and $\mathbf{0.731}$ on the private test set, providing a promising solution for identifying suicidal content in social media. Our analysis shows that the choice of LLMs affects the prompting performance, with larger models providing better accuracy. Our code and checkpoints are publicly available at \textcolor{pinkcolor2}{\textit{\href{https://github.com/khanhvynguyen/Suicide_Detection_LLMs}{\url{https://github.com/khanhvynguyen/Suicide_Detection_LLMs}}}}.
\end{abstract}

\begin{IEEEkeywords}
large language models, text classification, limited labels, semi-supervised learning, prompt engineering, suicide detection, social media analysis
\end{IEEEkeywords}

\section{Introduction}
\label{sec:inetroduction}

Suicide is a significant societal issue, with over $700{,}000$ people taking their own lives and many more attempting to do so\cite{WHO_suicide_rate}. Unfortunately, the prevalence of suicidal thoughts and attempts is increasing. Thus, early identification of suicidal thoughts is crucial for preventing serious consequences and providing timely support. Social media platforms have emerged as potential sources for detecting suicidal thoughts and attempts, as people often share their experiences or seek help on these platforms. However, the sheer volume of new posts daily makes it impractical for mental health professionals to review all of them and offer assistance or resources.

Deep learning techniques in natural language processing (NLP) have shown promise in detecting suicidal content in social media posts by effectively recognizing subtle patterns in the text data. The process typically involves three main tasks: collecting textual data from social media platforms, labeling the data, and building a deep learning-based classifier. The process of labeling is time-consuming and requires domain experts, which results in very limited annotated datasets. Additionally, social media posts may contain vague or implicit intent, necessitating a strong language understanding ability for accurate classification. These factors pose a challenge in developing an effective model for detecting suicide.

\begin{table*}[htbp]
\small
\caption{Definition of four suicide risk levels with increasing severe risk levels~\cite{li2022suicide}}  
\centering
\begin{tabular}{ll}
 \toprule
 \textbf{Category}  & \textbf{Definition} \\  
 \midrule
Indicator & The post content has no explicit expression
concerning suicide. \\ 
Ideation & The post content has explicit suicidal expression
but there is no plan to commit suicide. \\ 
Behaviour &  The post content has explicit suicidal expression
and a plan to commit suicide or self-harming
behaviours.\\
Attempt &  The post content has explicit expressions
concerning historic suicide attempts. \\ 
 \bottomrule
\end{tabular}

\label{table:suicide_risk_level_definition}
\end{table*}

\begin{table}[hbtp]
\small
\caption{Distribution of suicide risk level annotation of $500$ labeled posts in the training set}  
\centering
\begin{tabular}{lcc}
 \toprule
 \textbf{Class} & \textbf{Number of posts} & \textbf{Percent Proportion (\%)} \\
 \midrule
Indicator & 129 & 25.8 \\ 
Ideation & 190 & 38.0 \\ 
Behaviour & 140 & 28.0\\
Attempt & 41 & 8.2 \\ 
 \bottomrule
\end{tabular}

\label{table:data_distribution}
\end{table}

In this study, we investigate the use of Large Language Models (LLMs) to classify signs of suicide from user posts within a semi-supervised setup. We start by proposing a method for generating pseudo-labels for unlabeled data. We annotate user posts using LLMs (\eg, \textit{Qwen2-72B-Instruct}~\cite{yang2024qwen2}) using prompting. To minimize the noise in the labeling process, we fine-tune two more models (\textit{Llama3-8B}~\cite{dubey2024llama} and \textit{DepRoBERTa}~\cite{poswiata-perelkiewicz-2022-opi}) on a small set of annotated data to filter out unreliable labels. The unlabeled data with pseudo-labels is combined with the small set of labeled data to form a new training set. We then fine-tune some more LLMs (\eg., \textit{Llama3-8B}, \textit{Gemma2-9B}) on the newly formed dataset, and evaluate the effectiveness of the models in suicide classification. Finally, we combine these fine-tuned models, together with prompting LLMs, to create an ensemble for a more robust and performant suicide detector. We apply our approach to \textit{the Suicide Ideation Detection on Social Media Challenge}, a track in the IEEE Big Data 2024 Big Data Cup. Additionally, we provide analysis and discussion to gain insights into the results, such as how different LLMs affect the prompting performance, and choices of the loss function when fine-tuning. In summary, our contributions are:
\begin{itemize}
    \item Using Large Language Models (LLMs) with prompting to generate pseudo labels for unlabeled datasets, mitigating the issue of limited labeled data.
  \item Investigating current state-of-the-art text classification methods using LLMs for suicide detection.
    \item Experimenting with these approaches, using \textit{the Suicide Ideation Detection on Social Media Challenge} dataset~\cite{li2022suicide} to find a robust and performant model for the task.
  \item Conducting a comprehensive ablation study to clarify the effectiveness of our method.
\end{itemize}


\section{Datasets and metrics}
\label{sec:datasets_and_metrics}

\subsection{Datasets}
\label{subsec:datasets}

The data used in this study is from the Suicide Risk On Social Media Detection Challenge 2024 \cite{li2022suicide}. The dataset comprises a training set of $500$ labeled and $1{,}500$ unlabeled Reddit posts. To acquire $500$ annotated posts, the authors gathered $139{,}455$ posts from $76{,}186$ users between 01/01/2020 and 31/12/2021. Various pre-processing steps were undertaken, including the removal of user identity-related data (\eg, names, addresses, emails, and links) to protect privacy, as well as the elimination of overlapping posts and comments from the same user. For each user, their last posts were considered as a representation of their latest suicide ideation state. Such posts were denoted as \textit{targeted posts}. Subsequently, $500$ \textit{targeted posts} from $500$ random users were selected for annotation from the filtered dataset, which contained a total of $3{,}998$ posts from $1{,}791$ users. The annotation scheme for the suicide risk level label is depicted in Table~\ref{table:suicide_risk_level_definition}. Statistics for the four annotated suicide risk categories from the $500$ labeled posts are outlined in Table~\ref{table:data_distribution}. It is worth mentioning that the training dataset is imbalanced, with only $8$\% labeled as \textit{Attempt}. Most of the posts are short, typically less than $2{,}000$ words, as illustrated in Fig.~\ref{fig:distribution_num_words}. An example of a Reddit post in the training set is shown as follows.
\smallbreak
``\textit{I want to end it, I want to end it but I don't know how, when or anything else}''. (True label: \textit{Ideation})

\begin{figure*}[htbp]
\centering
\includegraphics[width=1\linewidth]{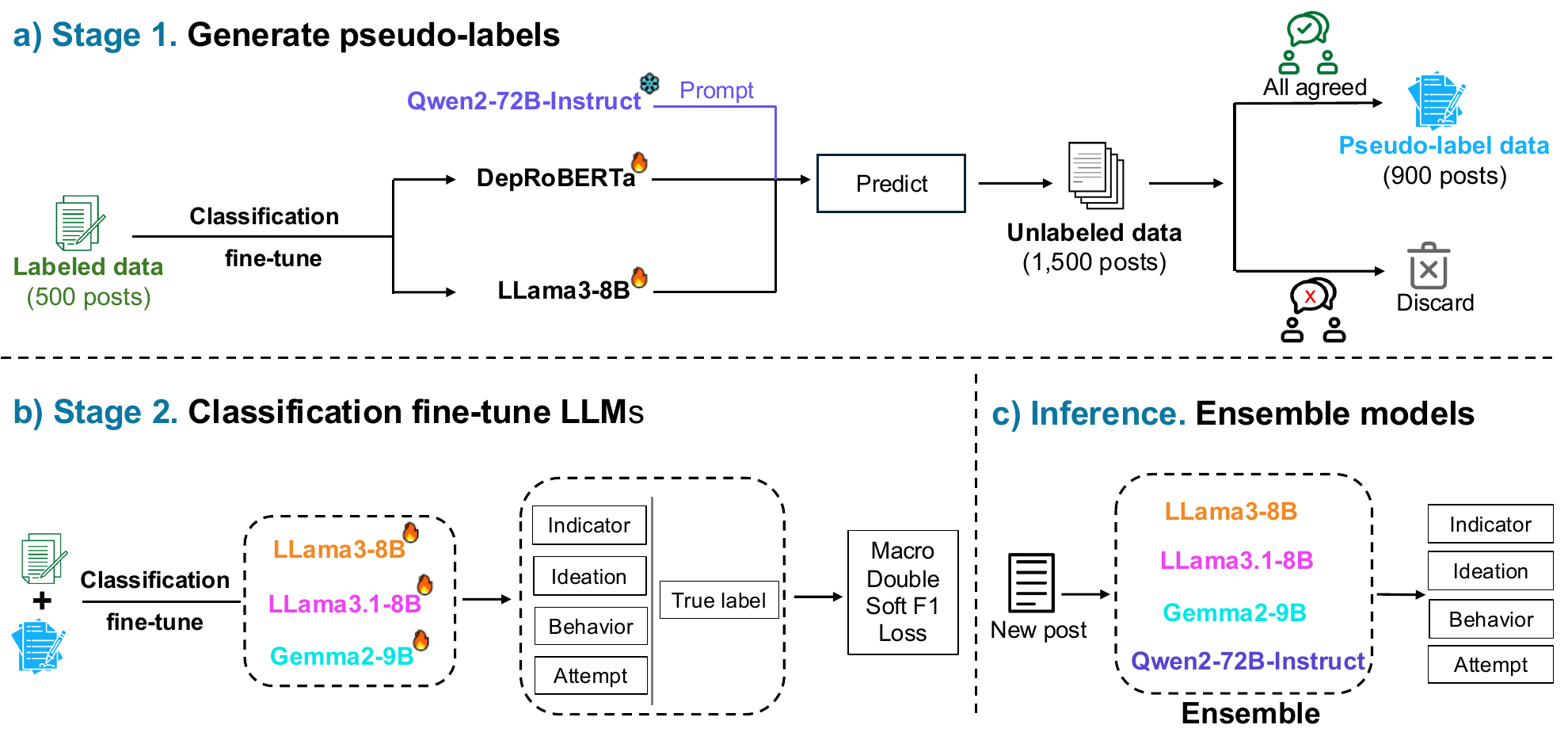}
\caption{\textbf{An overview of our approach.} \textbf{(a) pseudo-labels generation for unlabeled data}. We first use $500$ labeled posts to fine-tune \textit{DepRoBERTa}~\cite{poswiata-perelkiewicz-2022-opi} and \textit{Llama3-8B}~\cite{dubey2024llama} for the classification task. Then, we combine these models with \textbf{\textcolor{purplecolor}{\textit{Qwen2-72B-Instruct}}} via prompting to annotate $1{,}500$ posts in the unlabeled dataset. We keep only $\approx 900$ posts for which all three models predict the same and combine these with the $500$ labeled posts to form a new training set (Section \ref{subsec:generateing_pseudolabels}). \textbf{(b) LLMs fine-tuning}. We then fine-tune \textbf{\textcolor{orangecolor}{\textit{Llama3-8B}}}, \textbf{\textcolor{pinkcolor}{\textit{Llama3.1-8B}}}, and \textbf{\textcolor{sky}{\textit{Gemma2-9B}}} on the newly formed dataset with Macro Double Soft F1 loss (Section~\ref{subsec:fine_tune}). \textbf{(c) Model Ensembling}. These fine-tuned models are combined with prompting \textbf{\textcolor{purplecolor}{\textit{Qwen2-72B-Instruct}}} to create an ensemble model for classifying new user posts 
(Section~\ref{subsec:ensemble}).}
\label{fig:overview}
\end{figure*}

\subsection{Metrics}
\label{subsec:metrics}
 We report \textit{accuracy} and \textit{weighted F1} scores for evaluating the model's performance. Following prior work~\cite{li2022suicide}, we utilize the weighted F1 score as the main metric since it provides a balanced measure of precision and recall, while also addressing class imbalance as depicted in Table~\ref{table:data_distribution}. 
 
 Let $C$ represent the total number of classes, $N$ be the total number of observations, and $n_c$ be the number of observations belonging to class $c$. The weighted F1 score is defined as follows:

\begin{equation}
\mathrm{F1}_\text{weighted} = \sum_{c=1}^{C} \frac{n_c}{N} \cdot \mathrm{F1}_c
\end{equation}
where $\mathrm{F1}_c$ denotes the F1 score for class $c$. This method ensures a more accurate evaluation by reflecting the significance of each class in the overall performance metric. For simplicity, we use $\mathrm{F1}$ score to indicate weighted $\mathrm{F1}$ score in this paper.

\section{Related Work}
\label{sec:related_worl}
\subsection{Deep learning approaches for text classification}


In this section, we discuss three primary approaches to text classification: \textit{Feature extraction}, \textit{Classification fine-tuning}, and \textit{Large Language Models (LLMs) via prompt engineering}.

\smallbreak

\noindent\textbf{Feature extraction approaches} begin with extracting features from the input text and then training a classifier for the downstream task. Some traditional methods such as \textit{Word2Vec}~\cite{mikolov2013efficient} and \textit{GloVe}~\cite{pennington-etal-2014-glove} have been widely used. These methods represent words as high-dimensional vectors learned from word co-occurrence statistics. However, they often struggle to capture the complexities of language as they cannot capture the context in the input. Transformer-based models (\eg, \textit{BERT}~\cite{devlin-etal-2019-bert}, \textit{LLaMA3}~\cite{dubey2024llama}, \textit{GPT-4}~\cite{OpenAI2023}), have recently become more commonly used as feature extractors due to their high performance~\cite{muennighoff-etal-2023-mteb}. These models, which were trained on large corpora of text, use attention mechanisms~\cite{vaswani2017attention} to capture the context of words in a sentence. This enables them to understand the meaning of words in a specific context. The embeddings produced by the LLM's pretrained parameters serve as input features for training traditional machine learning classifiers like \textit{XGBoost}~\cite{chen2016xgboost}. By freezing the LLM's parameters during this process, we can leverage its ability to capture intricate linguistic patterns without the need for extensive fine-tuning. This hybrid approach combines the strengths of LLMs in feature extraction with the simplicity of traditional classifiers, offering an efficient method for text classification.

\smallbreak
\noindent\textbf{Classification fine-tuning} involves updating a model's parameters on a specific dataset for a particular task. This is different from the feature extraction approach, where frozen language models are used to generate features for separate classifiers. Fine-tuning usually involves initializing the model's weights with its pretrained weights. Subsequently, the final layer is replaced with a new randomly initialized linear layer (\ie, classifier head) that maps the feature dimension to the number of classes. The model is then fine-tuned on a labeled dataset relevant to the target classification task. This approach utilizes transfer learning, where the knowledge gained during pretraining on extensive text data is applied to the specific task at hand.
Recent advancements have shown that fine-tuning can significantly improve the model's ability to capture domain-specific language patterns, leading to enhanced accuracy and reliability in predictions~\cite{parthasarathy2024ultimate,howard-ruder-2018-universal,lee2020biobert}. Our approach involves fine-tuning pre-trained models on suicide datasets to better capture the nuances of suicidal thoughts, thus improving predictive performance. However, fine-tuning these large models can be challenging. To effectively fine-tune LLMs, efficient training techniques are necessary, such as Flash Attention~\cite{dao2022flashattention} for faster attention computation with reduced memory requirements, bottleneck adapter layers for fine-tuning~\cite{houlsby2019parameter}, LoRA~\cite{hu2022lora} for decreasing parameter count using low-rank matrices, and Bitsandbytes for quantization to lower memory usage during training. 

\smallbreak
\noindent\textbf{LLMs via prompt engineering} have become a critical method for improving the capabilities of LLMs (\eg, \textit{LLaMA3}~\cite{dubey2024llama}, \textit{GPT-4}~\cite{OpenAI2023}) across various downstream tasks~\cite{sahoo2024systematic}. In this approach, a \textit{prompt} - that is, a task-specific instruction along with some examples (few-shot) and a question - can be directly input into LLMs to obtain a response for the question. This technique enhances the model's effectiveness by utilizing in-context learning, without the need to modify the model parameters. To further enhance the reasoning ability of LLMs, Wang~\etal~\cite{wang2023selfconsistency} introduced Self-Consistency,  where a single LLM generates multiple responses to a given question and then uses majority voting to determine the final answer.
Wei~\etal~\cite{wei2022chain} introduced Chain-of-Thought (CoT), a method that employs a series of intermediate reasoning steps to reach the final answer. In another line of work, Du~\etal~\cite{du2023improving} organized multiple LLMs in a debate setting, where each LLM independently generates an initial response, and then communicates to reach a consensus on a final answer. Meanwhile, Pham~\etal~\cite{phamCIPHER2024} removed the token sampling step from LLMs, allowing LLMs to debate more effectively through the raw transformer output embeddings. Overall, these techniques have been effective in improving the performance of LLMs through the use of prompts in downstream tasks. In our approach, we utilize LLMs with few-shot CoT prompting to generate pseudo-labels for unlabeled data, aiding in the creation of a substantial training set for the fine-tuning phase.






\subsection{Large Language Models for Mental Health Detection}

Recent research has expanded the use of large language models (LLMs) for detecting mental health issues. Lan \etal proposed combining LLMs and traditional classifiers to integrate medical knowledge-guided features, achieving both high accuracy and explainability in depression detection~\cite{lan2024depression}. Poswiata \etal introduced fine-tuning \textit{RoBERTa}~\cite{liu2019roberta} on depression-related posts to achieve an effective depression detector~\cite{poswiata-perelkiewicz-2022-opi}. Yang \etal~\cite{yang2024mentallama} introduced \textit{MentalLLaMA} based on \textit{LLaMA2}~\cite{touvron2023llama}, for interpretable mental health analysis with instruction-following capability. Arcan \etal~\cite{arcan2024assessment} provided a comprehensive evaluation of \textit{LLaMA2} and \textit{ChatGPT-4} for mental health tasks, revealing both the potential and limitations of LLM-based methods in this field. There is another line of work in utilizing open-source LLMs to identify evidence in online posts that indicates the level of suicidal risks~\cite{chim2024overview}. As for suicide detection, Li \etal~\cite{li2022suicide} proposed a new benchmark, consisting of fine-grained annotations of suicidal posts with BERT-based models. Meanwhile, Qi \etal~\cite{qi2023evaluating} presented another benchmark for Chinese social media datasets, assessing the performance of traditional models and  Chinese LLMs such as \textit{Chinese-LLaMa-2-7B}~\cite{cui2023efficient}. In this paper, we focus on using state-of-the-art LLMs under a limited labeled data setting to create an effective and reliable classifier for identifying suicidal content in English social media posts.

\section{Leveraging Large Language Models for suicide classification with limited labels}
\label{sec:our_method}

Given a user post (\ie., text) $T$, our goal is to train a model that takes $T$ as input to classify the suicide risk levels. Our method involves using Large Language Models (LLMs) with few-shot Chain-of-Thought prompting~\cite{wei2022chain}, and classification fine-tuning, as shown in Fig.~\ref{fig:overview}. First, we generate pseudo-labels for unlabeled data to mitigate the issue of limited labeled data (Section~\ref{subsec:generateing_pseudolabels}). High-confidence pseudo-labels are retained and combined with the labeled data to create the training set. Next, we fine-tune some LLMs such as \textit{Llama3-8B}~\cite{dubey2024llama}, \textit{Gemma2-9B}~\cite{team2024gemma} using the new training set with Macro Double Soft F1 loss\cite{multi_label_soft_f1} (Section~\ref{subsec:fine_tune}). Finally, we create an ensemble model using \textit{Qwen2-72B-Instruct}~\cite{yang2024qwen2} through prompting, along with the classification fine-tuned LLMs on the new dataset, resulting in a robust and high-performing suicide classifier (Section~\ref{subsec:ensemble}).

\subsection{Generating pseudo-labels for unlabeled data}
\label{subsec:generateing_pseudolabels}
\begin{figure}[tbp]
\centering
\includegraphics[width=1\linewidth]{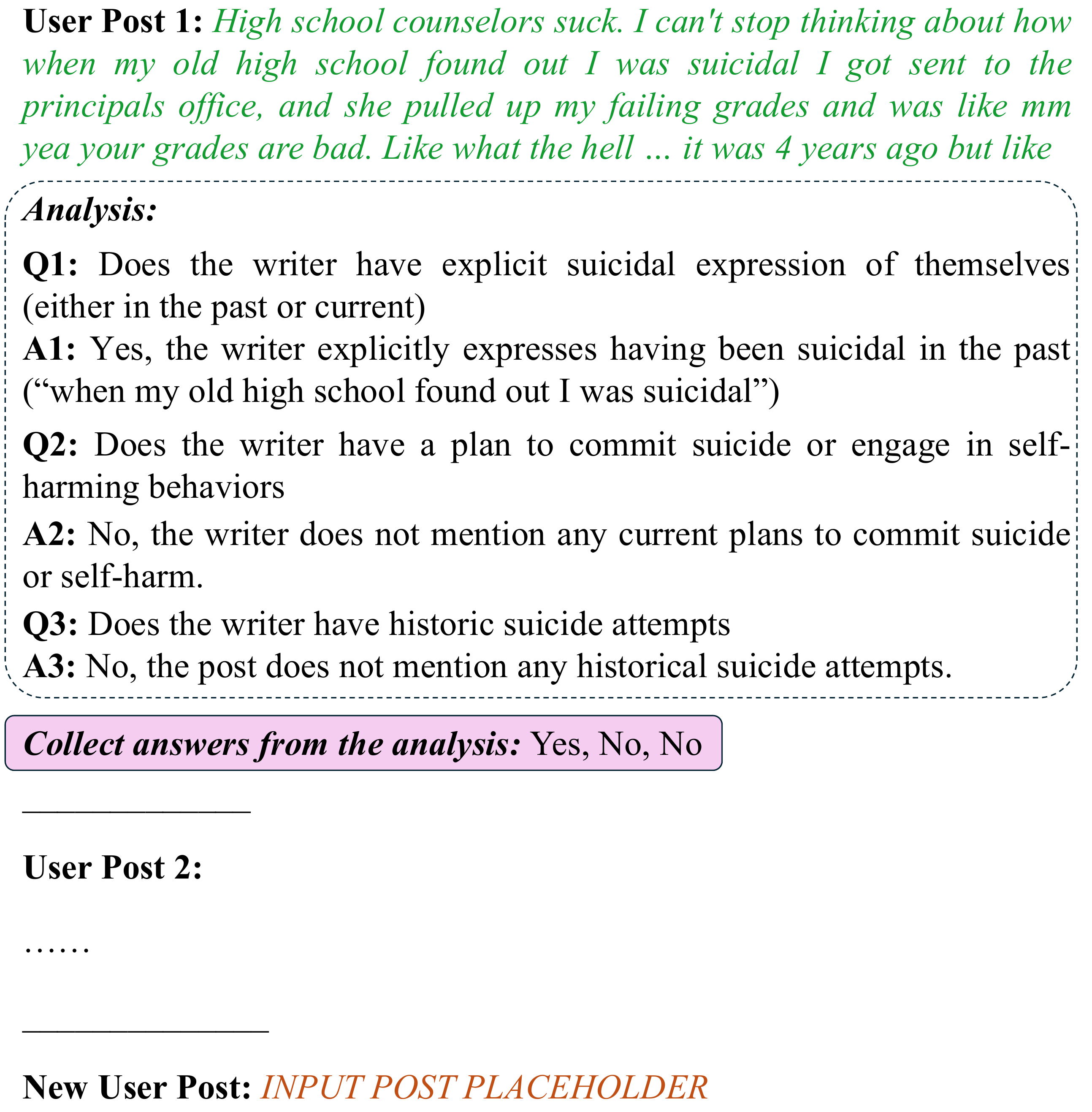}
\caption{\textbf{Prompt template with few-shot examples using Chain-of-Thought prompting.} Each example consists of a user post followed by three questions corresponding to the three suicide risk levels: \textit{Ideation}, \textit{Behavior}, and \textit{Attempt}. For each question in the example, a sample response is provided. The final part is used to collect the \textit{Yes/No} answers from the three responses.}
\label{fig:prompt}
\end{figure}

\noindent\textbf{Annotation with LLMs via prompting.}
Large language models (LLMs) have shown impressive semantic understanding capabilities~\cite{achiam2023gpt}. Research suggests they could potentially replace human annotators in some tasks~\cite{ziems-etal-2024-large, yuan-etal-2024-hide}. In this study, we leverage LLMs with promoting to generate pseudo-labels for $1{,}500$ posts in the unlabeled dataset. We manually composed a set of six few-shot examples with Chain of Thought (CoT) for prompting, demonstrated in Fig.~\ref{fig:prompt}. Each example includes a user post, highlighted in green, followed by three questions corresponding to the three suicide risk levels: \textit{Ideation}, \textit{Behavior}, and \textit{Attempt}. Each question is accompanied by a response, starting with a ``Yes" or ``No," followed by an explanation. The final part of the exemplar compiles these responses into a collective answer, consisting of the three ``Yes/No" answers from the questions.

When a new post replaces the orange-highlighted placeholder (Fig.~\ref{fig:prompt}), the LLM is expected to generate a response in the same format as the exemplar, including three questions, three corresponding answers, and a compiled final answer, all within the dashed box (Fig.~\ref{fig:prompt}).

 The final classification is determined by interpreting the compiled responses from the LLM, parsed in reverse order - from right to left. These answers correspond to \textit{Attempt}, \textit{Behavior}, and \textit{Ideation}, respectively. The classification is assigned based on the first ``Yes" encountered. For example, if the collected answer is \textit{\{Yes, Yes, No\}}, the post is labeled as \textit{Behavior}. If none of the responses in the collected answer is ``Yes," the post is classified as \textit{Indicator} (\ie, indicating no explicit expression regarding suicide). One advantage of using prompting LLMs is that the labels are interpretable by humans. By prompting the LLM to answer smaller questions, we can verify the reasoning behind the model's predictions. For example, in Fig.~\ref{fig:prompt}, the model believes that the writer has expressed explicit suicidal thoughts because of the phrase \textit{``when my old high school found out I was suicidal."}

\begin{figure}[tbp]
\centering
\includegraphics[width=1\linewidth]{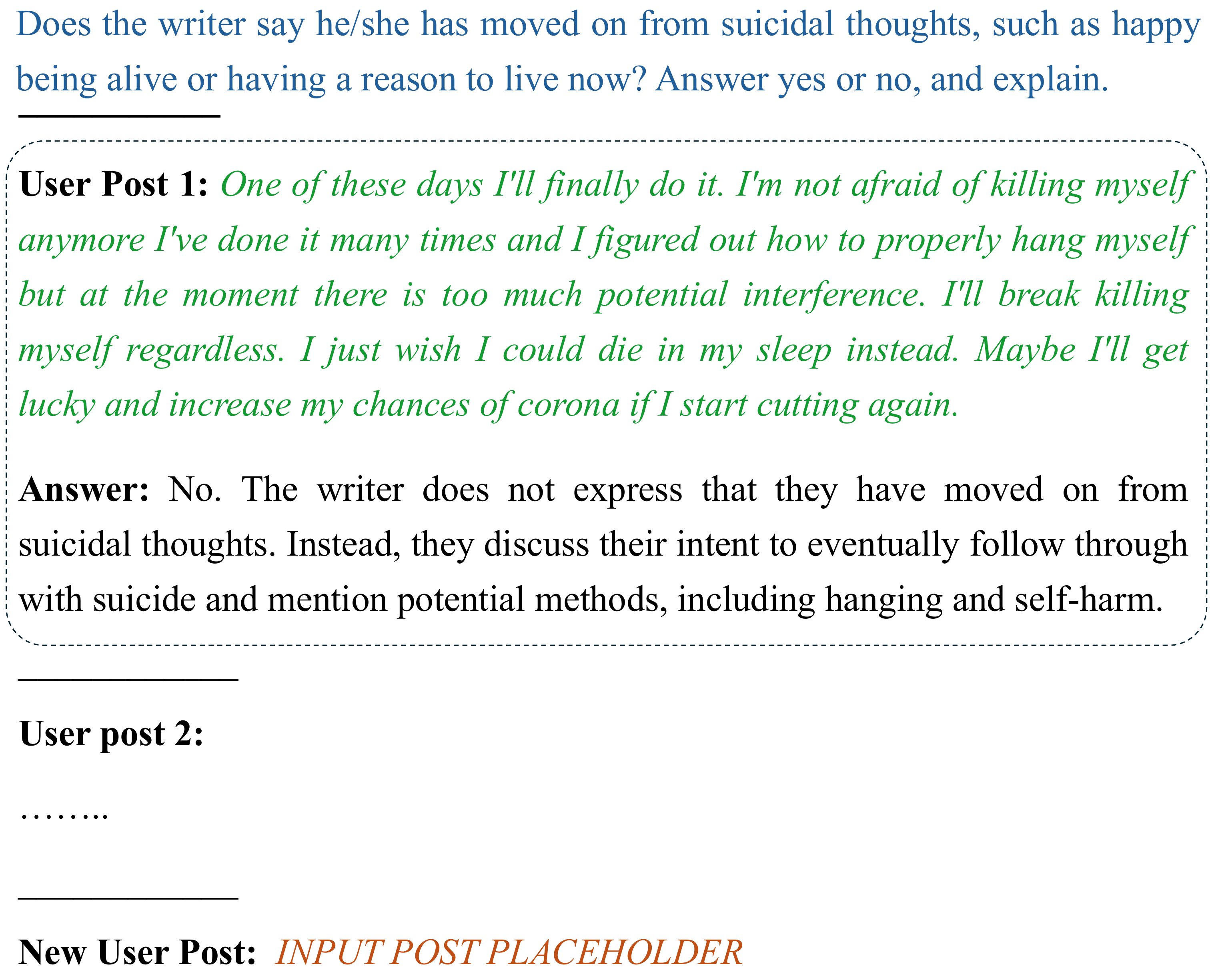}
\caption{\textbf{Prompt template to assess whether the writer mentions moving on from suicide thoughts or attempts.} First, an instruction is provided to guide the LLM on the task, highlighted in blue. Following this, a set of few-shot exemplars is presented. Each exemplar includes a user post followed by the corresponding answer. When a new post, highlighted in orange, replaces the placeholder, the LLM is expected to generate an answer based on the context of the post, indicating whether the writer mentions moving on.}

\label{fig:move_on}
\end{figure}

Through error analysis of the annotation results, we observed that some posts involved writers who had previously attempted suicide but then expressed a desire to continue living and move forward. LLMs may misinterpret the overall context of these posts, leading to incorrect annotations as \textit{Attempts}. To address this issue, we introduced an additional prompt specifically for posts initially labeled as \textit{Attempts} at the initial round. The prompt template is illustrated in Fig.~\ref{fig:move_on}. This prompt checks whether the writer mentions moving on from suicidal thoughts or attempts, such as feeling happy to be alive or having found a reason to live. If the answer generated by LLM is ``Yes," we modify the original answers by flipping the last answer corresponding to \textit{Attempt} to ``No." For example, if LLMs initially predict a post as \{\textit{Yes, No, Yes}\} and the writer mentions moving on, we change the last ``Yes" for \textit{Attempts} to ``No". The refined answers then become \textit{\{Yes, No, No\}}, and the post would be re-annotated as \textit{Ideation}. Another approach we tried was to label posts as \textit{Indicator} if the LLM initially predicted \textit{Attempt} but the writer displayed positive signs within the post. However, this method did not produce favorable results on the validation dataset.

We employ \textit{Qwen2-72B-Instruct}~\footnote{\url{https://huggingface.co/Qwen/Qwen2-72B-Instruct}} to generate pseudo-label for the unlabeled posts, as it is one of the performant open-source models with an extended context window of up to $128K$ tokens~\cite{yang2024qwen2}. We use greedy-decoding (\ie, choosing the highest probability output at each step of the sequence generation) and set the maximum length of the generated tokens to $1024$. We also experiment with other LLMs, such as \textit{LLaMA3}~\cite{dubey2024llama} and Mistral family models~\cite{jiang2023mistral,jiang2024mixtral}, refer to Table~\ref{tab:compare_llms} and Section~\ref{sec:ablation_study} for further discussion.


\smallbreak
\noindent\textbf{Annotation with classification fine-tuning.} To reduce noisy labels during the labeling process, we utilize multiple models for annotating the labels rather than relying solely on a single model. This ensures a more clean set of annotations, which is particularly important when dealing with complex and sensitive topics such as suicide risk.

To this end, we use $500$ labeled posts to fine-tune two different classes of transformer-based models: an Encoder-only model and a Decoder-only model. For the former, we utilize \textit{DepRoBERTa}, the winning solution for the Shared Task on Detecting Signs of Depression from Social Media Text at LT-EDI-ACL2022~\cite{poswiata-perelkiewicz-2022-opi}. Based on \textit{RoBERTa}~\cite{liu2019roberta}, \textit{DepRoBERTa} was further fine-tuned on user posts with different depression levels. Since depression and suicide are closely correlated, \textit{DepRoBERTa} serves as a promising starting point for the suicide prediction task. For the latter, we fine-tune Llama3-8B~\cite{dubey2024llama}. More details on how these models are fine-tuned are discussed in Section~\ref{subsec:fine_tune}.


\smallbreak
\noindent\textbf{Combining models to generate training dataset.} 
We use a combination of \textit{Qwen2-72B-Instruct}, along with fine-tuned \textit{DepRoBERTa} and \textit{Llama3-8B} models on the $500$ labeled posts to generate pseudo-labels for unlabeled data, as depicted in Fig.~\ref{fig:overview}(a), rightmost. Specifically, we only kept posts for which all three models predicted the same labels, and discarded the rest in case of disagreement. By doing so, we retain only high-confidence pseudo-labels, which allow us to reduce model-specific biases, resulting in a cleaner dataset for fine-tuning later on (refer to Table~\ref{table:data_cover} in Section~\ref{sec:ablation_study} for more discussion). After labeling $1{,}500$ posts, we keep only $902$ posts with pseudo-labels. These posts, combined with the $500$ labeled dataset, form the training set of $1{,}402$ posts. Statistics for the four annotated suicide risk labels from the combined dataset are outlined in Table~\ref{table:data_distribution_1402_new}.

\begin{table}[tbp]
\small
\caption{\textbf{Distribution of suicide risk level annotation of $\mathbf{1{,}402}$ labeled training set, consisting of $\mathbf{500}$ original labeled posts and $\mathbf{902}$ pseudo-labeled posts.}} 
\centering
\begin{tabular}{lcc} 
 \toprule
 \textbf{Class} & \textbf{Number of posts} & \textbf{Percent Proportion (\%)} \\  
 \midrule
Indicator & 540 & 38.5 \\ 
Ideation & 500 & 35.7 \\ 
Behaviour & 286 & 20.4 \\
Attempt & 76 & 5.4 \\ 
 \bottomrule
\end{tabular}

\label{table:data_distribution_1402_new}
\end{table}


\subsection{Fine-tuning Large Language Models for suicide classification}
\label{subsec:fine_tune}
After forming the new training set from Section~\ref{subsec:generateing_pseudolabels}, we fine-tune several LLMs on these $1{,}402$ posts. From Section~\ref{subsec:generateing_pseudolabels}, we observed that \textit{LLaMA3} outperformed \textit{DepRoBERTa} when fine-tuned on the $500$ labeled posts. As a result, we decide to focus on fine-tuning LLaMA~\cite{dubey2024llama} for this stage. To enhance the diversity of the results for the ensemble model later on, we also conduct fine-tuning on \textit{Gemma2}~\cite{team2024gemma}. Specifically, we fine-tune three different models: \textbf{\textcolor{orangecolor}{\textit{Llama3-8B}}}, \textbf{\textcolor{pinkcolor}{\textit{Llama3.1-8B}}}, and \textbf{\textcolor{sky}{\textit{Gemma2-9B}}}, as shown in Fig.~\ref{fig:overview}(b).

\smallbreak
\noindent \textbf{Setting for $\mathbf{5}$-fold cross-validation.} Although we have $1{,}402$ labeled posts available, we only use $500$ original labeled data annotated by experts for validation to ensure accuracy. To utilize all $500$ well-annotated posts, we use $5$-fold cross-validation training. When one fold is used for evaluation, the remaining four folds, along with $902$ pseudo-labeled posts, are combined for training. These folds are stratified to maintain the initial suicide risk level proportions. 

After training five cross-validation models, we collect the prediction for a new test set as follows. Let 
$c \in \{0,1,2,3\}$ represent the class index,
and $\mathbf{m^{(i)}} \in \mathbb{R}^4$ be the probability output of the $i$-th cross-validation model given a user post $T$. We average the probability outputs of all five models to obtain the final probability output $\mathbf{p} \in \mathbb{R}^4$ (E.q~\ref{eq:get_cv_prob}). Next, we select the class corresponding to the index with the highest probability, denoted as $\mathrm{c}\ast$ (E.q~\ref{eq:get_cv_pred}), as the final prediction.

\begin{equation}
\mathbf{p} =  \frac{1}{5} \sum_{i=1}^{5} \mathbf{m^{(i)}}
\label{eq:get_cv_prob}
\end{equation}

\begin{equation}
 \mathrm{c}\ast= \arg\max_{c} \mathbf{p_c} 
\label{eq:get_cv_pred}
\end{equation}

\smallbreak
\noindent \textbf{LoRA classification Fine-tuning.} In the fine-tuning process, the model's weights are initialized with its pretrained weights. After that, the final layer is replaced with a new linear layer that is randomly initialized and maps the feature dimension to the number of classes, which is 4 in this case. We fine-tune each model using a single NVIDIA RTX with 48GB RAM for $20$ epochs and a batch size of $1$. To account for the small batch size, we set grad accumulation to $16$. The AdamW optimizer~\cite{loshchilov2018decoupled} is used to train the models. For the learning rate, we apply a random search in the range of $1e\text{-}6$ to $1e\text{-}4$. Additionally, we use a weight decay of $0.1$ to the weight parameters to mitigate overfitting. To efficiently fine-tune large language models (LLMs), we employ LoRA~\cite{hu2022lora} to reduce the parameter count via low-rank matrices. The dimension of the low-rank matrices $\mathrm{lora}_r$  is set at $16$, whereas $\mathrm{lora}_ \alpha$ is set to $8$. We only apply LoRA fine-tune on $q_\mathrm{proj}, k_\mathrm{proj}, v_\mathrm{proj}$ and $o_\mathrm{proj}$ layers, with a dropout of $0.05$. Besides LoRA, we employ other efficient training techniques such as Flash Attention~\cite{dao2022flashattention} for faster attention computation with reduced memory requirements and Bitsandbytes~\footnote{\url{https://huggingface.co/docs/transformers/v4.44.2/quantization/bitsandbytes}} for quantization to decrease memory usage during training.

Due to computational constraints, we limit the maximum number of tokens for each post to $2{,}500$ tokens. This is chosen to cover the majority of posts without losing significant information, as demonstrated in Fig.~\ref{fig:distribution_num_tokens_llama3_8b} and Fig.~\ref{fig:distribution_num_tokens_gemma2-9b}. If a post exceeds this limit, we remove the middle portion, as prior work observed that the beginning and end of posts usually contain important information for text analysis~\cite{sun2019fine,garcia-etal-2023-deeplearningbrasil}. 

After fine-tuning for $20$ epochs, we select the model checkpoint with the highest F1 Score on the validation set as the final classifier.

\smallbreak
\noindent\textbf{Loss function.} Since F1 Score is computed via
 precision and recall, it is not differentiable. To directly optimize the F1 score, we use Macro Double Soft F1, introduced by~\cite{multi_label_soft_f1}, as our loss function. The idea is to make F1-score differentiable by modifying its computation. Specifically, \textit{True Positives, False Positives}, and \textit{False Negatives} are derived from a continuous sum of likelihood values using probabilities, eliminating the need for thresholds. We found that optimizing with this loss function in our experiment data gains some performance boost compared with other common choices such as Cross Entropy (Table~\ref{tab:compare_losses}). 
Algorithm~\ref{algo:macro_double_soft_f1} shows the pseudo-code of the loss.

\begin{algorithm}[tbp]
\footnotesize
\SetAlgoLined
\PyNormal{\textbf{Parameters}:}\\
\PyNormalTab{\textit{y}: one-hot vector of true labels, shape (batch size, num classes)} \\
\PyNormalTab{\textit{logits}: raw predictions, shape (batch size, num classes)} \\
\PyNormal{}\\
    \PyComment{Convert true labels to float} \\
    \PyCode{y = y.float()} \\
    \PyComment{Pass logits to sigmoid} \\

    \PyCode{y\_hat = F.sigmoid(logits.float())} \\
    \PyComment{Compute true positives, false positives, false negatives, and true negatives} \\
    \PyCode{tp = torch.sum(y\_hat * y, dim=0)} \\
    \PyCode{fp = torch.sum(y\_hat * (1 - y), dim=0)} \\
    \PyCode{fn = torch.sum((1 - y\_hat) * y, dim=0)} \\
    \PyCode{tn = torch.sum((1 - y\_hat) * (1 - y), dim=0)} \\
    \PyComment{Calculate soft F1 scores for class 1 and class 0, small epsilon prevents divided by zero} \\
    \PyCode{soft\_f1\_class1 = 2 * tp / (2 * tp + fn + fp + 1e-16)} \\
    \PyCode{soft\_f1\_class0 = 2 * tn / (2 * tn + fn + fp + 1e-16)} \\
    \PyComment{Calculate losses for both classes} \\
    \PyCode{cost\_class1 = 1 - soft\_f1\_class1} \\
    \PyCode{cost\_class0 = 1 - soft\_f1\_class0} \\
    \PyComment{Compute loss} \\
    \PyCode{cost = 0.5 * (cost\_class1 + cost\_class0)} \\
    \PyCode{soft\_f1\_loss = torch.mean(cost)} \\
    \PyComment{Return the f1 loss} \\
    \PyCode{\textbf{Return } soft\_f1\_loss}
\caption{Pseudo code of Macro Double Soft F1 Loss~\cite{multi_label_soft_f1} in Pytorch}
\label{algo:macro_double_soft_f1}
\end{algorithm}



\subsection{Ensemble Model for a robust and performant classifier}
\label{subsec:ensemble}

Ensemble modeling has been shown to significantly improve the accuracy and reliability of predictions (\eg, ~\cite{9006542,poswiata-perelkiewicz-2022-opi,9377988,garcia-etal-2023-deeplearningbrasil}). In the same spirit, we create an ensemble model by using majority voting of class predictions from some individual models.
Let $\mathrm{C}$ be the set of all classes ($\mathrm{C}=$\{\textit{Indicator}, \textit{Ideation}, \textit{Behaviour}, \textit{Attempt}\}), $\mathrm{m}_i$ be the class prediction of model $i$ ($\mathrm{m}_i \in \mathrm{C}$), $\mathrm{w}_i \in \mathbb{R}$ be the ensemble weight of model $i$, and $n$ denote the number of models to ensemble. The weighted majority voting $\mathrm{p}$ over all prediction $\mathrm{m}_i$ is defined as follows:

\begin{equation}
\arg \max _{\mathrm{p}} \sum_{i=1}^n \mathrm{w}_i \cdot \mathbb{1}  \left(\mathrm{m}_i=\mathrm{p}\right) 
\label{eq:ensemble}
\end{equation}

Our ensemble model consisting of five individual models: \textbf{\textcolor{purplecolor}{\textit{Qwen2-72B-Instruct}}} via prompting, and four fine-tuned large language models (LLMs) - \textbf{\textcolor{orangecolor}{\textit{Llama3-8B}}} (with two variants differing in hyperparameters, denoted as \textit{Llama3-8B 1} and \textit{Llama3-8B 2}), \textbf{\textcolor{pinkcolor}{\textit{Llama3.1-8B}}}, and \textbf{\textcolor{sky}{\textit{Gemma2-9B}}}, as shown in Fig.~\ref{fig:overview}(c). From $500$ labeled data, we assign the ensemble weight $\mathrm{w}_i$ to the models, with a weight of $2$ for \textit{Qwen2-72B-Instruct}, and a weight of $1$ for the other four models. This means that the prediction of \textit{Qwen2-72B-Instruct} counts twice, whereas all other models count once. The final prediction $\mathrm{p}$ is determined through a weighted majority voting over the five models as illustrated in Eq.~\ref{eq:ensemble}. This ensemble approach combines the strengths of each individual model to create a more performant and robust classifier, as demonstrated and discussed in Section~\ref{sec:results} and Section~\ref{sec:model_evaluation}.

\section{Results}
\label{sec:results}

\begin{figure}[tbp]
\centering
\includegraphics[width=0.8\linewidth]{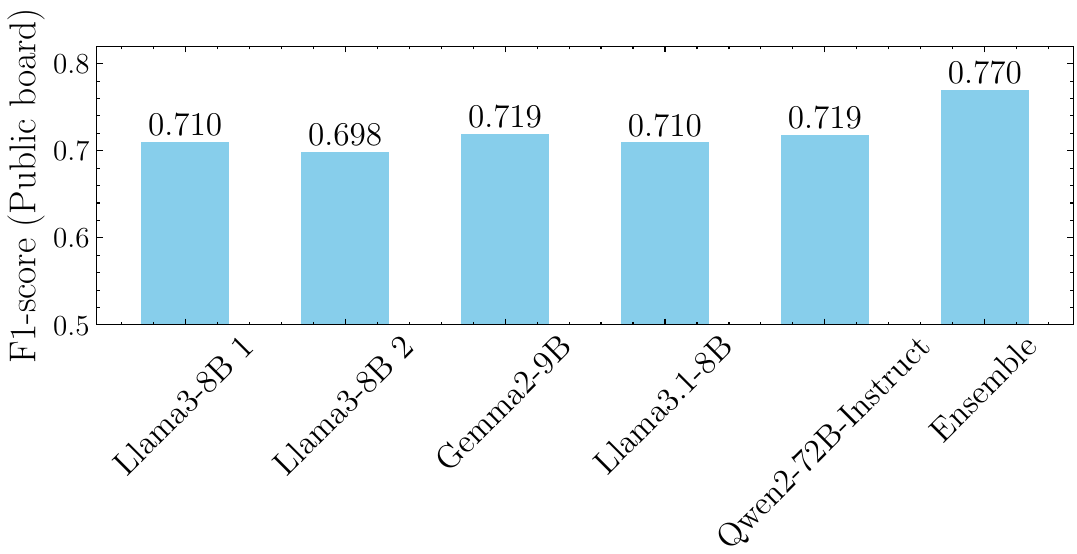}
\caption{\textbf{Comparison of F1 Scores for our models on the Public Board Test Set containing $\mathbf{100}$ posts.} The ensemble model shows its robustness and significantly outperforms the individual models, demonstrating an improvement of approximately $5$\% points in the F1 Score on the new test set.} 
\label{fig:model_f1_testset}
\end{figure}

\begin{table}[tbp]
\small
\caption{\textbf{Comparison of Accuracy and F1 Scores for our models in the $\mathbf{5}-$fold cross-validation.} The ensemble model outperformed the individual models in both accuracy and F1 score, with improvements of around $1-4$\% in accuracy and $0.01-0.04$ in the F1 Score.} 

\centering

\begin{tabular}{lcc} 
 \toprule
 \textbf{Model} & \textbf{Accuracy (\%)} & \textbf{F1 Score} \\ 
 \midrule
Llama3-8B 1 & 77.0 $\pm$ 2.3 & 0.762 $\pm$ 0.030 \\ 
Llama3-8B 2 & 77.4 $\pm$ 3.1 & 0.771 $\pm$ 0.032 \\ 
Gemma2-9B & 80.6 $\pm$ 2.1 & 0.805 $\pm$ 0.023 \\ 
Llama3.1-8B & 79.0 $\pm$ 2.4 & 0.789 $\pm$ 0.025 \\ 
Qwen2-72B-Instruct & 77.4 $\pm$ 2.9 & 0.772 $\pm$  0.029 \\ 
\textbf{Ensemble} & \textbf{81.2 $\pm$ 2.3} & \textbf{0.811 $\pm$ 0.023} \\ 
 \bottomrule
\end{tabular}
\label{tab:acc_f1_scores}
\end{table}

Table \ref{tab:acc_f1_scores} presents a comparison of the average accuracy and F1 scores, along with their standard deviations, for our models using $5$-fold cross-validation. We can observe the ensemble model outperforms the five individual models. Specifically, \textit{Llama3-8B} and \textit{Qwen2-72B-Instruct} achieve approximately $77$\% accuracy, while \textit{Gemma2-9B} and \textit{LLaMA3.1-8B} score $79-80$\% accuracy. The ensemble model achieves the highest accuracy at $81.2$\%. Similarly, the ensemble model also attained the highest F1 score of $0.811$, compared to $0.76-0.80$ for the individual models.

Fig.~\ref{fig:model_f1_testset} presents the comparison of F1 Scores for our models on the test set from the Public Board of the competition. It is noticeable that the F1 scores in the test set which consists of $100$ new user posts, are lower than those in the validation set. This discrepancy may be attributed to the differences in label distribution between the two datasets. That said, the ensemble model demonstrates strong robustness and still performs well on the test set. Specifically, the ensemble model achieves the highest F1 score on the test set at $0.770$, significantly better than any individual model by $5$\% points.

\section{Analysis and Discussion}
\label{sec:analysis_and_discussion}

\subsection{Ablation Study: loss functions, model size for prompting, and noise reduction for pseudo-labels}
\label{sec:ablation_study}

\noindent\textbf{Using model agreement in Stage 1 reduces the noise in the labeling process.} Table~\ref{table:data_cover} shows the accuracy of each model in \textit{Stage 1} when trained on $5$-fold cross-validation with $500$ labeled posts. Each individual model achieved an accuracy below $78$\%. Combining the agreement of all three models boosts the accuracy to $88.5$\%, at the cost of smaller data coverage, with only $64.6$\% of the posts being predicted. This demonstrates the effectiveness of using three models to reduce noise when generating pseudo-labels, illustrated in Fig.~\ref{fig:overview}(a).

\smallbreak
\noindent\textbf{Ablation on Loss functions}. Table~\ref{tab:compare_losses} compares the accuracy and F1 scores obtained from 5-fold cross-validation when training\textit{ Llama3-8B} (\textit{Stage 2)} using different loss functions. The results show that Macro Double Soft F1 outperforms Cross Entropy in accuracy and F1 score in this experiment. Specifically, Macro Double Soft F1 achieves a higher accuracy of $77.4$\% and an F1 score of $0.77$1, while these metrics are lower for Cross Entropy, at $76.0$\% and $0.760$, respectively.

\begin{table}[tbp]
\small
\caption{\textbf{Accuracy of each model in Stage 1 when trained on $\mathbf{5}$-fold cross-validation on $\mathbf{500}$ labeled posts.} The combined agreement of all models (\textit{All agreement)} results in a higher accuracy of $88.5$\%, showing its effectiveness in reducing the noise in the labeling process, at the cost of reduced data coverage of $64.6$\%.}
\centering
\begin{tabular}{lcc} 
 \toprule
 \textbf{Model} & \textbf{Accuracy (\%)} & \textbf{Data Cover (\%) } \\  
 \midrule
Llama3-8B & 74.9 & 100.0 \\ 
DepRoBERTa & 69.6 & 100.0 \\ 
Qwen2-72B-Instruct & 77.4 & 100.0 \\
\textbf{All agreement} & \textbf{88.5} & \textbf{64.6} \\ 
 \bottomrule
\end{tabular}

\label{table:data_cover}
\end{table}

\begin{table}[tbp]
\small
\caption{\textbf{Comparison of different choices of loss functions.} We report accuracy and F1 Scores on 5-fold cross-validation when training \textit{Llama3-8B} (\textit{Stage 2}).} 
\centering
\begin{tabular}{lcc} 
 \toprule
 \textbf{Model} &  \textbf{Accuracy} (\%) & \textbf{F1 Score}  \\
 \midrule
\textbf{Macro Double Soft F1}~\cite{multi_label_soft_f1} & \textbf{77.4 $\pm$ 3.1} & \textbf{0.771 $\pm$ 0.032}  \\ 
Cross Entropy & 76.0 $\pm$ 2.7  & 0.760 $\pm$ 0.029 \\ 
 \bottomrule
\end{tabular}%
\label{tab:compare_losses}
\vspace{4mm}
\end{table}

\begin{table}[t!]
\small
\caption{\textbf{Comparison of LLMs with prompting on 500 labeled posts}. \textit{Qwen2-72B-Instruct}, the largest model evaluated, outperforms the smaller models.} 
\centering
\begin{tabular}{lcc} 
 \toprule
 \textbf{Model} &  \textbf{Accuracy} (\%) & \textbf{F1 Score}  \\
 \midrule
Qwen2-7B-Instruct~\cite{yang2024qwen2}  & 50.0 & 0.469 \\
Mistral-7B-Instruct-v0.3~\cite{jiang2023mistral}  & 57.0 & 0.591 \\
Llama3-8B~\cite{dubey2024llama} & 48.0 & 0.419 \\
Llama3-8B-Instruct~\cite{dubey2024llama} & 55.0 & 0.503 \\
Mixtral-8x22B~\cite{jiang2024mixtral}  & 60.0 & 0.594 \\
Llama3-70B-Instruct~\cite{dubey2024llama}  & 69.4 & 0.693  \\ 
Llama3-70B~\cite{dubey2024llama}  & 74.2 & 0.741   \\ 
\textbf{Qwen2-72B-Instruct}~\cite{yang2024qwen2} &\textbf{77.4 }& \textbf{0.772}  \\ 

 \bottomrule
\end{tabular}
\label{tab:compare_llms}
\end{table}

\begin{figure*}[t]
\centering
\begin{subfigure}[t]{0.22\textwidth}
    \centering
    \includegraphics[width=\textwidth]{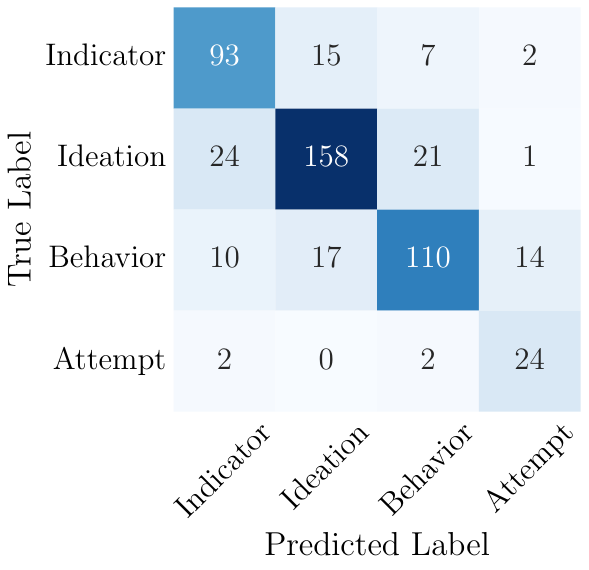}
    \caption{Llama3-8B 1 }
    \label{fig:l8b_1}
\end{subfigure}
\qquad
\begin{subfigure}[t]{0.22\textwidth}
    \centering
\includegraphics[width=\textwidth]{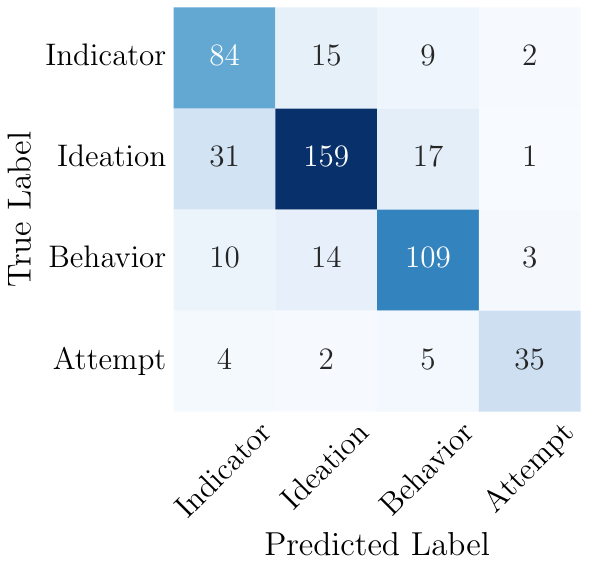}
    \caption{Llama3-8B 2}
    \label{fig:l8b_2}
\end{subfigure}
\qquad
\begin{subfigure}[t]{0.22\textwidth}
    \centering
    \includegraphics[width=\textwidth]{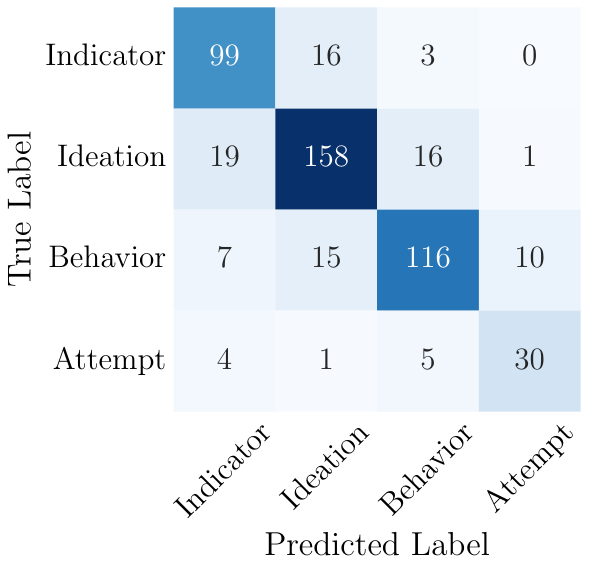}
    \caption{Gemma2-9B}
    \label{fig:gi}
\end{subfigure}
\smallbreak
\smallbreak
\smallbreak

\begin{subfigure}[t]{0.22\textwidth}
    \centering
    \includegraphics[width=\textwidth]{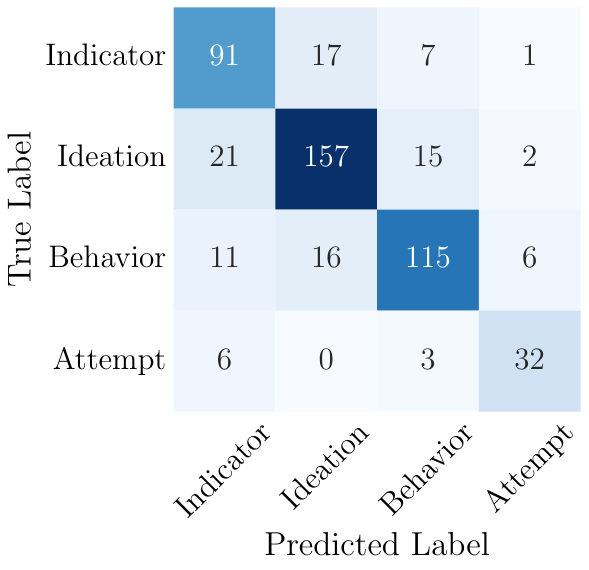}
    \caption{Llama3.1-8B}
    \label{fig:l31_8b}
\end{subfigure}
\qquad
\begin{subfigure}[t]{0.22\textwidth}
    \centering
    \includegraphics[width=\textwidth]{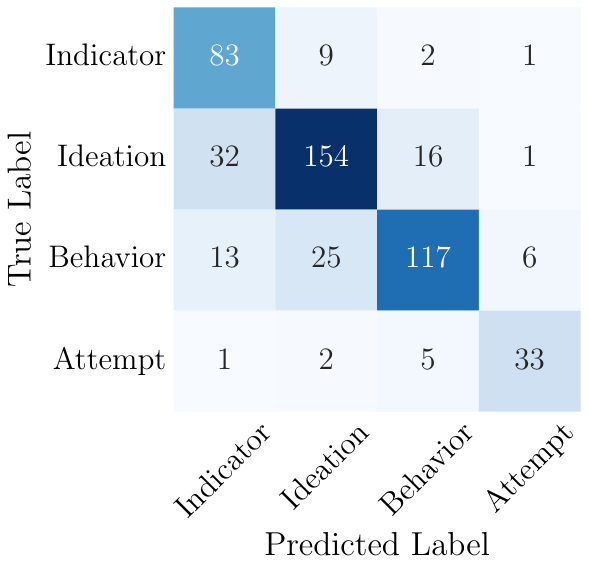}
    \caption{Qwen2-72B}
    \label{fig:qwen}
\end{subfigure}
\qquad
\begin{subfigure}[t]{0.22\textwidth}
    \centering
    \includegraphics[width=\textwidth]{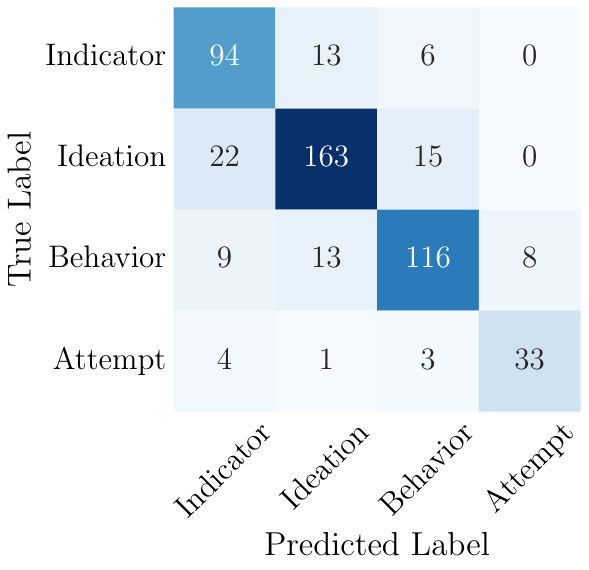}
    \caption{Ensemble}
    \label{fig:ensemble}
\end{subfigure}
\caption{\textbf{Confusion Matrices of the models on 500 original labeled posts.} The matrices show that each of the models performs reasonably well. \textit{Ensemble} (f) outperforms individual models, with the main improvement coming from \textit{Ideation} class.}
\label{fig:model_confusion_matrix}
\end{figure*}

\begin{table}[btp]
\small
\centering
\caption{\textbf{Classification Report of the ensemble model.} The model achieves a weighted F1 Score of $0.81$, indicating consistent performance across all classes}
\begin{tabular}{lccc}
\toprule
     \textbf{Class}      & \textbf{Precision} & \textbf{Recall} & \textbf{F1 Score} \\ 
      \midrule
Indicator & 0.83 & 0.73 & 0.78 \\
Ideation & 0.81 & 0.86 & 0.84 \\
Behavior & 0.79 & 0.83 & 0.81 \\
Attempts & 0.80 & 0.80 & 0.80 \\ 
 \midrule
{Accuracy} & &   & 0.81 \\
{Macro avg} & 0.81 & 0.81 & 0.81 \\
{Weighted avg} & 0.81 & 0.81 & \textbf{0.81 }\\ 
\bottomrule
\end{tabular}
\label{tab:classification_report}
\end{table}

\begin{figure}[tbp]
\centering
\includegraphics[width=0.75\linewidth]{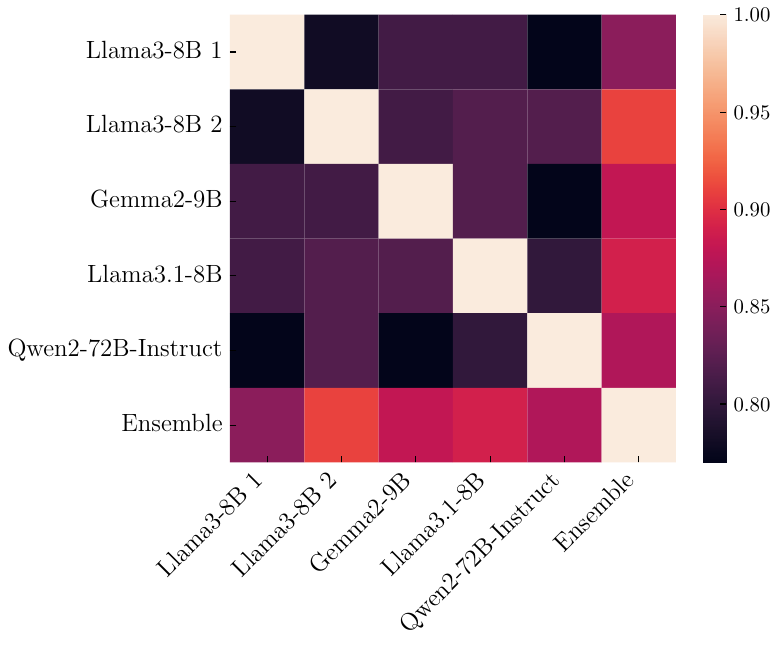}
\caption{\textbf{Correlation Between Model Predictions on 500 labeled data.} The ensemble model demonstrates a stronger correlation with each model (last row and column), when compared to the pair of individual models.}
\label{fig:model_heatmap}
\end{figure}

\smallbreak
\noindent\textbf{Impact of different LLMs on prompting performance.} Table~\ref{tab:compare_llms} presents a comparison of the accuracy and F1 scores across LLMs with prompting on $500$ labeled posts. We observe that larger models generally achieve higher performance metrics. Specifically, the largest model evaluated, \textit{Qwen2-72B-Instruct}, with $72$ billion parameters, attains the highest accuracy of $77.4$\% and an F1 score of $0.772$, demonstrating superior understanding and classification capability compared to smaller models. Following closely are \textit{Llama3-70B} and \textit{Llama3-70B-Instruct}, with accuracies of $74.2$\% and $69.4$\% and F1 scores of $0.741$ and $0.693$, respectively. Conversely, smaller models such as \textit{Qwen2-7B-Instruct} and \textit{Llama3-8B} demonstrate lower performance. Among relatively small models, \textit{Mistral-7B-Instruct-V03} exhibits the highest performance, nearly on par with \textit{Mixtral-8x22B}, a mixture of eight experts with $39$B active parameters out of $141$B parameters.

\begin{figure*}[btp]
\centering
\begin{subfigure}[t]{0.32\textwidth}
\centering
    \includegraphics[width=\textwidth]{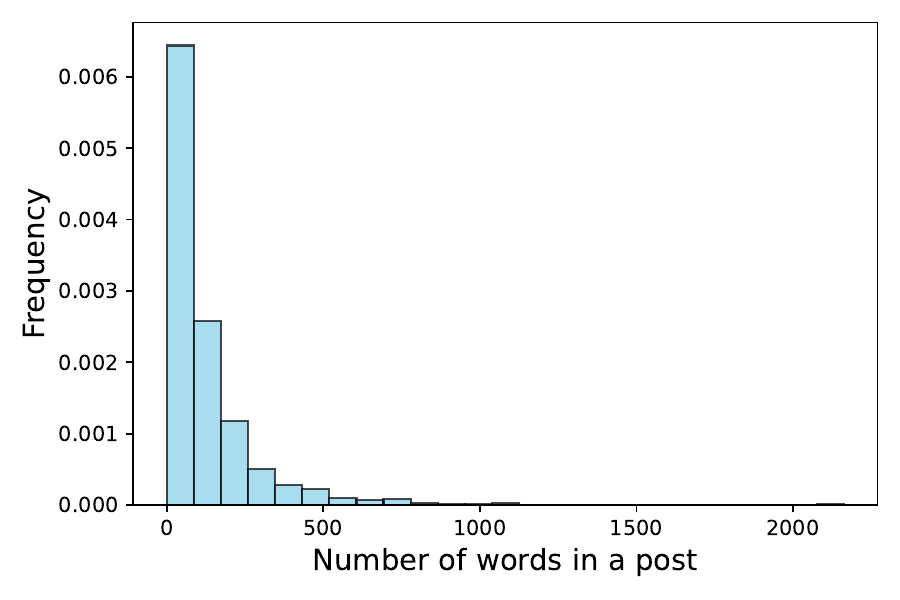}
    \caption{\textbf{Number of words}}
    \label{fig:distribution_num_words}
\end{subfigure}
\hfill
\begin{subfigure}[t]{0.32\textwidth}
\centering
    \includegraphics[width=\textwidth]{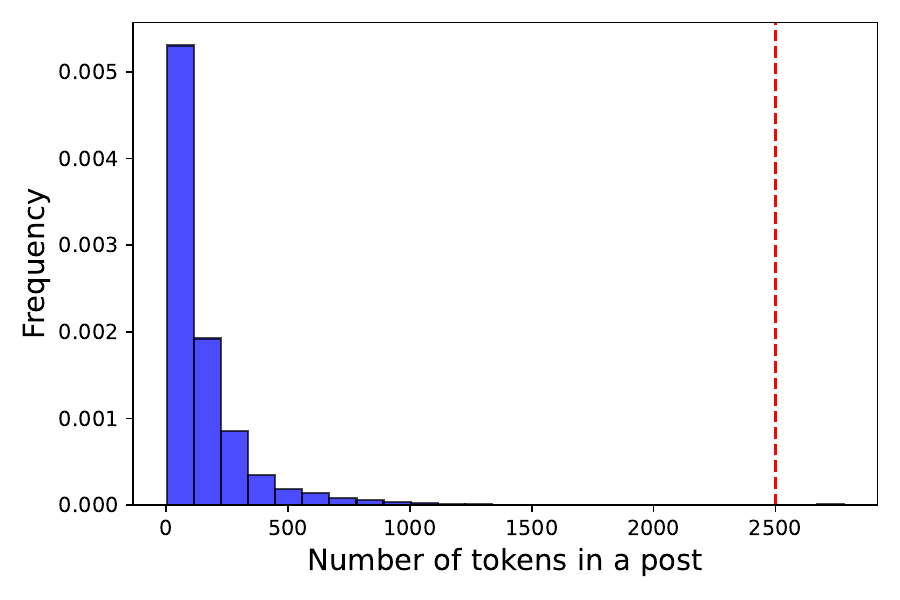}
    \caption{\textbf{Llama3-8B}}
    \label{fig:distribution_num_tokens_llama3_8b}
\end{subfigure}
\hfill
\begin{subfigure}[t]{0.32\textwidth}
\centering
    \includegraphics[width=\textwidth]{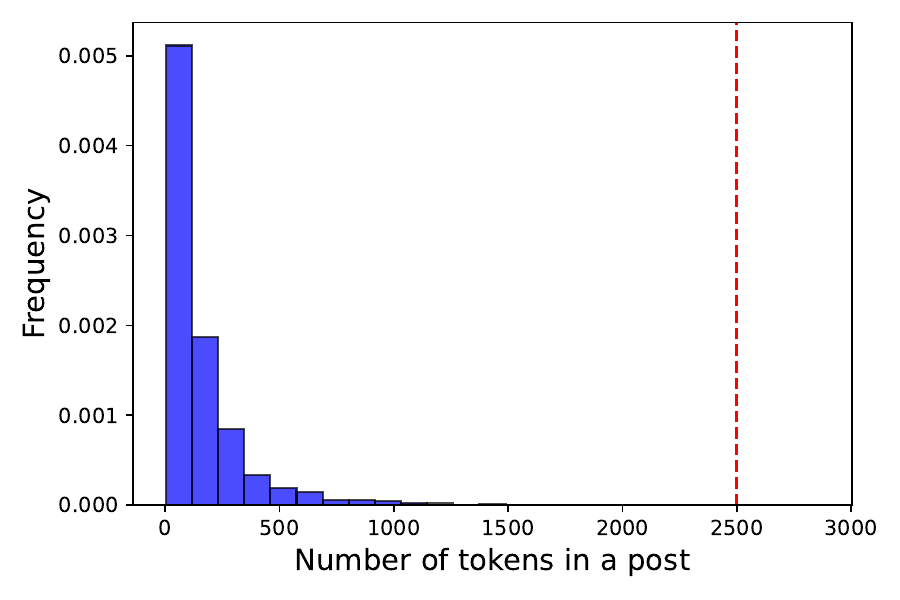}
    \caption{\textbf{Gemma2-9B}}
    \label{fig:distribution_num_tokens_gemma2-9b}
\end{subfigure}
\caption{\textbf{Distribution of the number of words/tokens in a post for $\mathbf{2{,}000}$ posts in the dataset.} \textbf{(a) Distribution of the number of words in a post.} Most posts have fewer than 2000 words. \textbf{(b) and (c) Distribution of the number of tokens in a post for the Llama-8B and Gemma2-9B models.} In both models, only $2$ out of $2{,}000$ posts exceed $2{,}500$ tokens. The red dashed line represents the $2{,}500$-token cut-off threshold.}
\label{fig:distribution_3figs}
\end{figure*}


\subsection{Model Evaluation and Performance}
\label{sec:model_evaluation}

\smallbreak
\noindent \textbf{Confusion Matrices of the classification results}. Fig.~\ref{fig:model_confusion_matrix} presents the confusion matrices for the individual and ensemble models' classification of the $500$ labeled posts. We observe  each of the individual models performs reasonably well. However, ensemble model outperforms individual models, achieving accuracies of $72.9$\%, $85.6$\%, $82.9$\%, and $80.5$\% for the \textit{Indicator}, \textit{Ideation}, \textit{Behavior}, and \textit{Attempt} categories, respectively, with the main improvement coming from \textit{Ideation} class. Despite these improvements, the ensemble model struggles with the \textit{Indicator} category, which has the highest rate of misclassification. Specifically, $22$ out of $129$ \textit{Indicator} posts were incorrectly classified as \textit{Ideation}, likely due to an over-sensitivity to certain features in the individual models.

\smallbreak
\noindent \textbf{Classification Report of ensemble model}.
Table~\ref{tab:classification_report} shows the precision, recall, and F1 Score of the ensemble model on the $500$ labeled posts. We can observe that the \textit{Ideation} category has the highest recall, at $0.86$. Both \textit{Behavior} and \textit{Attempts} show balanced precision and recall, with both around $0.80$. Overall, the model achieves a weighted F1 Score of $0.81$, indicating consistent performance across all classes.

\subsection{Prediction Correlation and Token Distribution Analysis}
\label{sec:correlation_and_token}
\smallbreak
\noindent\textbf{Correlation Between Model Predictions.} As shown in Fig.~\ref{fig:model_f1_testset}, the ensemble model demonstrates its robustness when evaluated on a new dataset. We compare the correlations between the predictions of the models, as illustrated in Fig.~\ref{fig:model_heatmap}. Our observations indicate that the ensemble model exhibits a stronger correlation with each individual model compared to the correlations between pairs of individual models.
\smallbreak
\noindent \textbf{Distributions of the number of tokens in a post.} Fig.~\ref{fig:distribution_3figs}(a) shows the distribution of the number of words in a post for $2{,}000$ training data samples. We can see that most posts contain fewer than $2{,}000$ words. Fig.~\ref{fig:distribution_3figs}(b) and Fig.~\ref{fig:distribution_3figs}(c) present the distribution of the number of tokens in a post for \textit{Llama3-8B} and \textit{Gemma2-9B}, respectively. The figures reveal that the majority of posts have fewer than $2{,}500$ tokens. In both models, only $2$ out of the $2{,}000$ posts exceed $2{,}500$ tokens.

\smallbreak
\noindent Among $47$ registered teams, our solution achieved a score of $75.436$ - the highest score on overall evaluation, based on \textit{model performance} ($6$-th on the private board containing a new set of $100$ user posts, with an F1 score of $0.7312$), \textit{approach innovation}, and \textit{report quality}. 

\section{Conclusion}
\label{sec:conclusion}
This study introduces a new method for identifying suicidal content on social media by using Large Language Models (LLMs) along with traditional fine-tuning techniques. By creating pseudo-labels through LLM prompting and fine-tuning models, we tackle the issue of having limited annotated datasets. We then build an ensemble method, utilizing models like prompting \textit{Qwen2-72B-Instruct}, and fine-tuned \textit{Llama3-8B} and \textit{Gemma2-9B}, which shows significant improvements in detection accuracy and robustness. The results from the Suicide Ideation Detection on Social Media Challenge validate the effectiveness of our approach, providing a promising solution for early suicide risk identification on social platforms. 

However, our work has some limitations. The ensemble model, particularly \textit{Qwen2-72B-Instruct}, requires a substantial amount of time for inference due to its large size, which poses challenges for real-time deployment. To mitigate this, smaller distillation models~\cite{muralidharan2024compact} could offer a more practical solution. Additionally, enhancing suicide detection using LLMs would benefit from better prompt engineering by domain experts. Breaking prompts into smaller, more specific questions can help direct LLMs to produce more relevant, accurate, and interpretable responses. Another promising approach is to collect more unlabeled data, leading to additional pseudo-labels for more effective fine-tuning of LLMs. Furthermore, incorporating visual data, such as images and videos from user posts, represents a promising direction for future research and may necessitate the use of multimodal LLMs.

\bibliographystyle{./IEEEtran}
\bibliography{my_bib} 

\end{document}